\documentclass[a4paper]{rnti}
\usepackage[T1]{fontenc}
\usepackage[utf8]{inputenc}
\usepackage{graphicx}
\usepackage{float}
\usepackage{hyperref}
\usepackage{listings}
\usepackage[caption=false]{subfig}
\usepackage{amsmath,bm,amsfonts,amssymb}
\usepackage{algorithm}
\usepackage{algorithmic}
\usepackage{xcolor} 
\usepackage[export]{adjustbox}
\usepackage{amsthm}

\titrecourt{Modèle de co-clustering de données mixtes}

\titre{Un modèle Bayésien de co-clustering de données mixtes}

\auteur{Aichetou Bouchareb\affil{1}\affilsep\affil{2}, Marc Boullé\affil{1}, Fabrice Rossi\affil{2}, Fabrice Clérot\affil{1}}
\affiliation{
    \affil{1}Orange Labs :\\
    prenom.nom@orange.com\\
    \affil{2}SAMM EA 4534 - 
    Université Paris 1 Panthéon-Sorbonne : \\ 
          prenom.nom@univ-paris1.fr
}
\nomcourt{A. Bouchareb et al.}
\resume{
Nous proposons un modèle de co-clustering de données mixtes et un critère Bayésien de sélection du meilleur modèle. Le modèle  infère automatiquement les discrétisations optimales de toutes les variables et effectue un co-clustering en minimisant un critère Bayésien de sélection de modèle.   Un  avantage de cette approche est qu'elle ne nécessite aucun paramètre utilisateur. De plus, le critère proposé  mesure de façon exacte la qualité d'un modèle tout en étant régularisé. L'optimisation de ce critère permet donc d'améliorer continuellement les modèles trouvés sans pour autant sur-apprendre les données.  Les expériences réalisées sur des données réelles montrent l'intérêt de cette approche pour l'analyse exploratoire des grandes bases de données.
}

\summary{ 
We propose a MAP Bayesian approach to perform and evaluate a co-clustering of  mixed-type data tables. The proposed model infers an optimal segmentation of all variables then performs a co-clustering by minimizing a Bayesian model selection cost function.  One advantage of this approach is that it is user parameter-free. Another main advantage is the proposed criterion which gives an exact measure of the model quality, measured by probability of fitting it to the data. Continuous optimization of this criterion ensures finding better and better models while avoiding data over-fitting. The experiments conducted on real data show the interest of this co-clustering approach in exploratory data analysis of large data sets. 
}

\begin{document}
\section{Introduction}
Dans un monde où les technologies d'acquisition de données sont en croissance rapide, l'analyse exploratoire des bases de données hétérogènes  et de grandes tailles  reste un domaine peu étudié. Une technique fondamentale de l'analyse non supervisée est celle du clustering, dont l'objectif est de découvrir la structure sous-jacente des données en regroupant les individus \emph{similaires} dans des groupes homogènes. Cependant, dans de nombreux contextes d'analyse exploratoire de données, cette technique de regroupement d'objets reste insuffisante pour découvrir les motifs les plus pertinents. Le co-clustering \citep{hartigan1975}, apparu comme extension du clustering, est une technique non-supervisée dont l'objectif est regrouper conjointement les deux dimensions de la même table de données, en profitant de l'interdépendance entre les deux entités (individus et variables) représentées par ces deux dimensions pour extraire la structure sous-jacente des données. Cette technique est la plus adaptée, par example, dans des contextes comme l'analyse des paniers de consommation où l'objectif est d'identifier les sous-ensembles de clients ayant tendance à acheter les mêmes de produits, plutôt que de grouper simplement les clients (ou les produits) en fonction des modèles d'achat/vente.

Dans la littérature, plusieurs approches de co-clustering ont été développées.  
 En particulier, certains algorithmes de co-clustering proposent d'optimiser une fonction qui mesure l'écart entre la matrice de données et la matrice de co-clusters \citep{church2000}. 
 D'autres techniques sont basées sur la théorie de l'information  (\cite{dhillon2003}), sur les modèles de mélange pour définir des modèles de blocs latents \citep{govaert2008}, sur l'estimation Bayésienne des paramètres  (\cite{banerjee2008}), sur l'approximation matricielle \citep{seung2001}, 
 ou sur le partitionnement des graphes \citep{dhillon2001}. 
 Cependant, ces méthodes s'appliquent naturellement sur des données de même type.

Dans \cite{bouchareb2017}, nous avons proposé une méthodologie permettant d'étendre l'utilisation du co-clustering au cas d'une table de données contenant des variables numériques et catégorielles simultanément. L'approche est basée sur une discrétisation de toutes les variables en fréquences égales, suivant un paramètre utilisateur, suivi par  l'application d'une méthode de co-clustering sur les données discretisées. Dans ce papier, nous proposons une nouvelle famille de modèles permettant de formaliser cette méthodologie. Le modèle proposé ici ne nécessite aucun paramètre utilisateur et permet une inférence automatique des discrétisations optimales des variables  selon une approche regularisée, par opposition à la discrétisation définie par l'utilisateur proposée par \cite{bouchareb2017}.   Un nouveau critère, mesurant la capacité du modèle à représenter les données, et de nouveaux algorithmes sont présentés.

Le reste de ce papier est organisé comme suit. En section~\ref{ModelCriterion}, nous présentons le modèle proposé, le critère de sélection et la stratégie d'optimisation implémentée. La section~\ref{Experiments}  présente des résultats expérimentaux sur des données réelles,  et la section~\ref{Conclusion} conclusion et perspectives. 
\section{Un modèle de co-clustering de données mixtes}\label{ModelCriterion}
Avant de présenter le modèle proposé, décrivons les données telles qu'elles sont vues par notre modèle. Les données sont composées d'un ensemble d'instances (identifiants de ligne de la matrice) et un ensemble de variables pouvant être numériques ou catégorielles. Nous définissons la notion d'une observation qui représente un 'log' d'une interaction entre  une instance et une variable. Cette représentation nous permet de considérer le cas des valeurs manquantes dans les données mais aussi le cas de plusieurs observations par couple (instance, variable) comme dans les séries temporelles.  Un example simple, illustrant cette représentation, est donné par :
\begin{center}
\footnotesize{
\begin{tabular}{cc}
\begin{tabular}{c}
\\
$i_1\rightarrow$\\
$i_2 \rightarrow$\\
$i_3\rightarrow$\\
$i_4\rightarrow$\\
\end{tabular} & \begin{tabular}{c}
\begin{tabular}{ccccc}
$X_1$ & $X_2\;\;\;$ & $X_3$ & $\;\;X_4$ & $\;\;X_5$ \\
\end{tabular}\\
$\begin{bmatrix} 
0 & -1 & .& \{b, a\} & A   \\
3&   \{0.2, 1, 0\} & 0 & b & B   \\
2  & . & 5 & \{a, c\} & A   \\
. & 1  &  22 & c & C  \\
\end{bmatrix}
$
\end{tabular}
\end{tabular}
}
\end{center}
Cet  example contient  $4$ instances ($i_1,\ldots,\,i_4$), 3  variables numériques ($X_1, X_2, X_3$), 2  variables catégorielles ($X_4, X_5$) et un total de $21$ observations.  

\subsection{Les paramètres du modèle}
Le modèle de co-clustering est défini par une hiérarchie des paramètres. A chaque étage de la hiérarchie, les paramètres sont choisis en fonction des paramètres précédents.

\newtheorem{definition}{Définition}
\begin{definition}
\label{definitionModel}
Le modèle de co-clustering des données mixtes est défini par : 
\begin{itemize}
	\item la taille de la partition de chaque variable. Une partition est un regroupement des valeurs dans le cas d'une variable catégorielle et une discrétisation en intervalles dans le cas d'une variable numérique,
	\item la partition des valeurs de chaque variable catégorielle en groupes de valeurs, 
	\item le nombre de clusters d'instances et de clusters de parties de variables. Ces choix définissent la taille de la matrice des co-clusters,
		\item la partition des instances et des parties de variables selon le nombre de clusters choisi, 
	\item la distribution des observations sur les cellules de la matrice des co-clusters, 
	\item la distribution des observations associées à chaque cluster d'instances  (resp. parties de variables)  sur l'ensemble des instances (resp.  parties de variables) dans le cluster,
	\item la distribution des observations dans chaque partie de variable catégorielle  sur l'ensemble des valeurs dans la partie. 
\end{itemize}
\end{definition}
\paragraph{Notations.}
Pour formaliser ce modèle, nous considérons les notations suivantes :
\begin{itemize}
\item[$\bullet$] $N$: le nombre total d'observations (connu),  
\item[$\bullet$] $K_n$: le nombre de variables numériques (connu),
\item[$\bullet$] $K_c$: le nombre de variables catégorielles (connu). $\mathbf{X}_c$  l'ensemble de ces variables,
\item[$\bullet$] $V_k$: le nombre de valeurs uniques de la variable catégorielle $X_k$ (connu),
\item[$\bullet$] $J_k$: le nombre de parties de la variable $X_k$ (\textbf{inconnu}),
\item[$\bullet$] $I$: le nombre total d'instances (connu),
\item[$\bullet$] $J=\sum_k{J_k}$: le nombre total de parties de variables (déduit),
\item[$\bullet$] $G_u$ : le nombre de clusters d'instances (\textbf{inconnu}),
\item[$\bullet$] $G_p$ : le nombre de clusters de parties de variables (\textbf{inconnu}),
\item[$\bullet$] $G = G_u\times G_p$ : le nombre de co-clusters (déduit),
\item[$\bullet$] $N_{g_u, g_p}$ : le nombre d'observations  dans le co-cluster formé par le cluster d'instances  $g_u$  et le  cluster de parties de variables  $g_p$ (\textbf{inconnu}),
\item[$\bullet$] $N^{(u)}_{g_u}$ : le nombre d'observations dans le cluster d'instances $g_u$ (déduit),
\item[$\bullet$] $N^{(p)}_{g_p}$ : nombre d'observations  dans le cluster de parties de variables  $g_p$ (déduit),
\item[$\bullet$] $m^{(u)}_{g_u}$ : le nombre d'instances dans le  cluster d'instances $g_u$ (déduit),
\item[$\bullet$] $m^{(p)}_{g_p}$ : le nombre de parties dans le  cluster de parties de variables $g_p$ (déduit),
\item[$\bullet$] $m^{(k)}_{j_k}$ : le nombre de valeurs dans la partie $j_k$ de la variable $X_k$ (déduit)
\item[$\bullet$] $n_{i.}$ : le nombre d'observations associées à la  $i^{\grave{e}me}$ instance (\textbf{inconnu}),
\item[$\bullet$] $n_{.kj_k}$ : le nombre d'observations associées à la partie $j_k$ de la variable $X_k$ (\textbf{inconnu})

\item[$\bullet$] $n_{v_{k}}$ : le nombre d'observations associées à la  valeur $v_k$ de la variable catégorielle $X_k$ (\textbf{inconnu})
\end{itemize}

Un modèle de la définition~\ref{definitionModel} est complètement défini par le choix des paramètres ci-dessus notés \textbf{inconnu}.
\subsection{Le critère Bayésien de sélection du meilleur modèle}
Nous faisons l'hypothèse d'une distribution a priori des paramètres la moins informative possible, en exploitant la hiérarchie des paramètres avec un a priori uniforme à chaque niveau.

Étant donné les paramètres, la vraisemblance conditionnelle $P(\mathcal{D}|\mathcal{M})$  des données sachant le modèle peut être définie par une distribution multinomiale sur chaque niveau de la hiérarchie. Le produit de la probabilité a priori du modèle et de la vraisemblance, permet de calculer de manière exacte la probabilité a posteriori du modèle connaissant les données $P(\mathcal{M}|\mathcal{D})$.  A partir de cette probabilité, nous définissons un critère de sélection de modèle $\mathcal{C}(\mathcal{M}) = -\log P(\mathcal{M}|\mathcal{D})$, donné par théorème \ref{criterion}.

\newtheorem{theorem}{Théorème}
\begin{theorem}\label{criterion}
Parmi les modèles  définis en définition~\ref{definitionModel}, un modèle suivant un a priori hiérarchique uniforme est optimal s'il minimise le critère  :

{\footnotesize	
\begin{equation}\label{CriterionEquation}
\begin{split}
\mathcal{C}(\mathcal{M}) =&  \sum\limits_{X_k\in \mathbf{X}_c}\log V_k +K_n \log N+\sum\limits_{X_k\in \mathbf{X}_c}{\log B(V_k, J_k)} +\log I +\log J \\
&  + \log B(I, G_u)+ \log B(J, G_p)+ \log \binom{
N+G-1}{G-1}+  \sum\limits_{g_u=1}^{G_u}\log \binom{
N_{g_u}^{(u)}+m_{g_u}^{(u)}-1}{m_{g_u}^{(u)}-1}\\
& +\sum\limits_{g_p=1}^{G_p} \log \binom{
N_{g_p}^{(p)}+m_{g_p}^{(p)}-1}{m_{g_p}^{(p)}-1} + \sum\limits_{X_k\in \mathbf{X}_c}\sum\limits_{j_k=1}^{J_k} \log \binom{
n_{.kj_k}+m_{j_k}^{(k)}-1}{m_{j_k}^{(k)}-1} \\
& + \log N! -\sum\limits_{g_u=1}^{G_u} 
  \sum\limits_{g_p=1}^{G_p} {\log
  N_{g_u, g_p}!}  + \sum\limits_{g_u=1}^{G_u}{\log N_{g_u}^{(u)}!} - \sum\limits_{i=1}^{I}{\log  
  n_{i.}! } \\
  &
 +\sum\limits_{g_p=1}^{G_p}{\log N_{g_p}^{(p)}! } -  \sum\limits_{X_k\in \mathbf{X}_c} 
  \sum\limits_{v_k =1}^{V_k}{\log 
  n_{v_k}!}
\end{split} 
\end{equation}
}

où $B(A, B)=\sum\limits_{b=1}^{B}{S(A, b)}$ est le nombre de Stirling de deuxième espèce donnant le nombre de répartitions possibles de $A$ valeurs en, au plus, $B$ groupes. 
\end{theorem}
Les trois première lignes représentent le coût a priori du modèle tandis que les deux dernières représentent le coût de la vraisemblance.  Pour des raisons de manque d'espace, la preuve de ce théorème n'est pas présentée dans ce papier.

\subsection{Algorithme d'optimisation}\label{OptimizationStrategy}
En raison de leur grande expressivité, les modelés de co-clustering des données mixtes sont complexes à optimiser. Dans ce papier, nous proposons une heuristique d'optimisation en deux étapes. Dans la première étape,  nous commençons par partitionner les variables en fréquences  égales en utilisant un ensemble prédéfini des tailles de partitions et nous appliquons la méthodologie proposée en \cite{bouchareb2017} pour trouver des co-clusters initiaux. Parmi les tailles testées, nous choisissons la solution initiale qui correspond à la valeur minimale du critère~ \eqref{CriterionEquation} comme point de départ. A partir de cette solution initiale, la deuxième étape est une post-optimisation qui effectue les fusions de clusters, les fusions de parties de variables, les déplacements de parties de variables entre clusters et déplacements de valeurs entre parties, qui minimisent le mieux le critère. Cette post-optimisation permet de choisir le meilleur modèle parmi un large sous-ensemble de modèles testés tout en améliorant l'interpretabilité, étant donné que le modèle optimisé est souvent très compact, comparé à la solution initiale. 
\section{Expérimentation}\label{Experiments}
Pour valider l'apport du modèle proposé dans l'analyse exploratoire des données mixtes, nous l'avons appliqué sur les bases de données Iris et CensusIncome \citep{datasets}.  

La base Iris est composée de  $150$ instances, $750$ observations, $4$ variables numériques et $1$ variable catégorielle. Les tailles des partitions de départ sont de 2 à 10 parties par variable. 

\begin{figure}[h]
\centering
\subfloat[Iris : le meilleur modèle]{\includegraphics[width=0.4\textwidth , height=3.6cm]{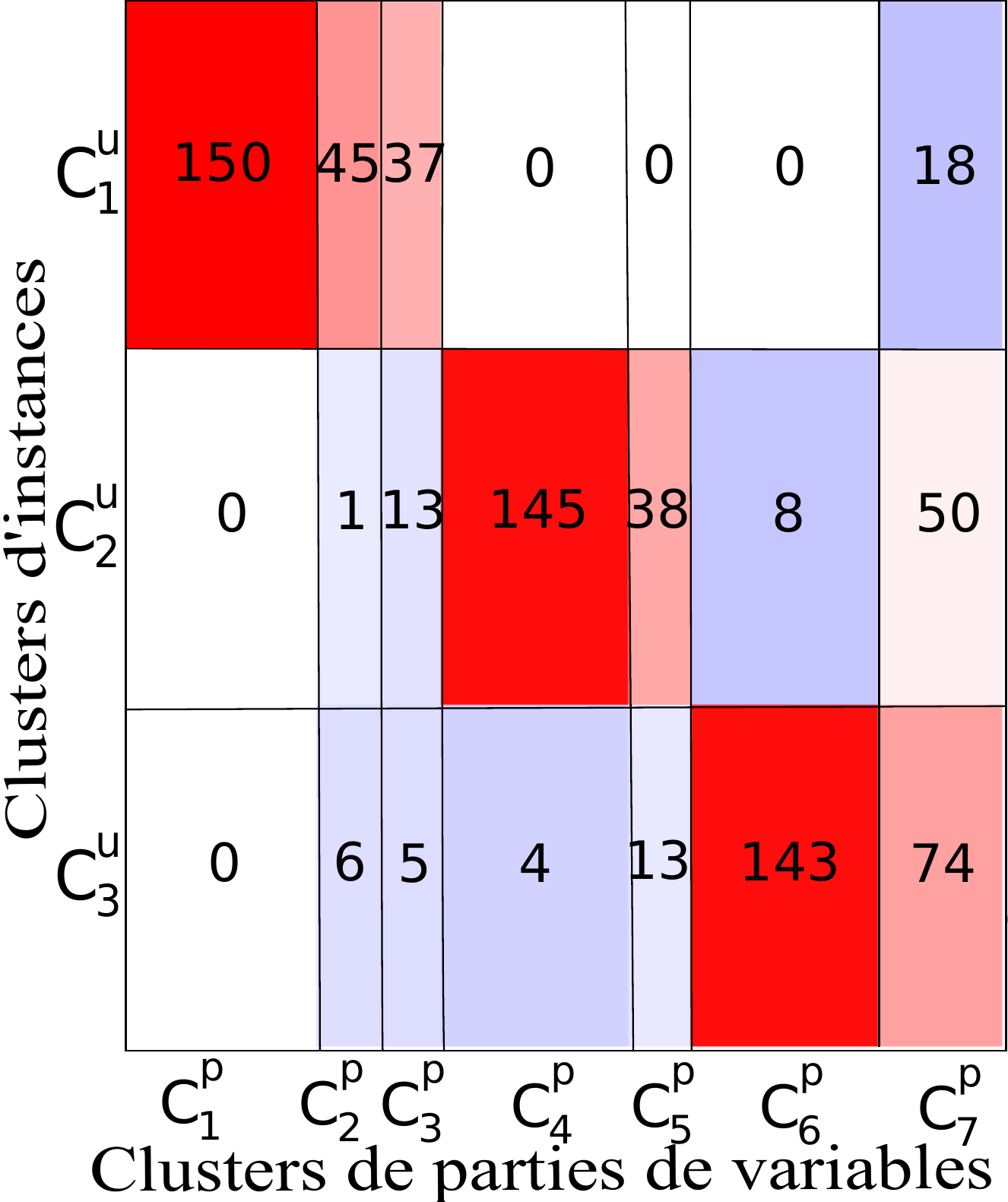}\label{IrisCoclusters}}
\quad
\subfloat[CensusIncome : $2\times 12$ Co-clusters]{    \includegraphics[width=0.4\textwidth, height=3.6cm]{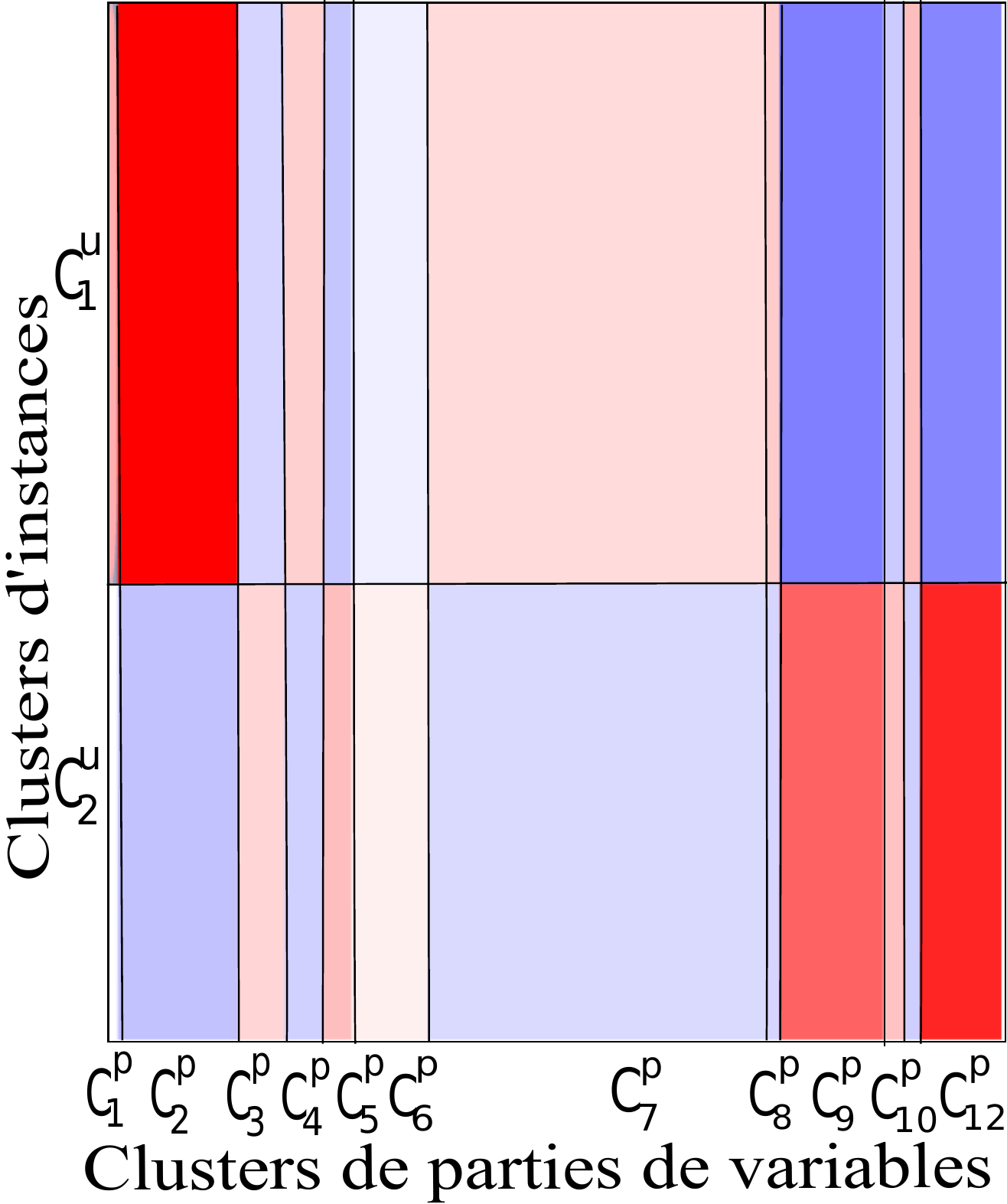}\label{CoclustersCensus2}}
\caption{{\footnotesize (a) : le meilleur modèle représentant la base Iris. (b) : un modèle simplifié de la base CensusIncome.}}
\label{Iris}
\end{figure}

La figure~\ref{IrisCoclusters} montre le meilleur modèle pour la base Iris. Ce modèle est le résultat d'une discrétisation initiale en $3$ parties par variable en fréquences égales suivi d'une post-optimisation qui fusionne deux parties pour en faire $14$ au total.  La couleur du co-cluster montre l'information mutuelle entre les instances et les parties de variables formant le co-cluster. La couleur rouge représente une sur-représentation des observations par rapport au cas d'indépendance. La couleur bleu représente une sous-représentation et la couleur blanche un co-cluster vide. Pour confirmation, le nombre d'observations par co-cluster est montré sur la figure~\ref{IrisCoclusters}.  
 
 Le modèle optimisé comporte 3 clusters d'instances et 7 clusters de parties de variables. Les compositions des clusters de parties de variables les mieux représentés permettent d'expliquer les clusters d'instances. En particulier, nous distinguons :
\begin{itemize}
\item un cluster ($C_1^u$) de 50 instances contenant les  petites fleurs  setosa caractérisées par $C_1^p$ (i.e.  \emph{Class}$\{setosa\}$, \emph{PetalLength}$]-inf;2.4]$, et \emph{PetalWidth}$]-inf;0.8]$),
\item  un cluster ($C_2^u$) de 51 instances contenant les grandes fleurs virginica caractérisées par  $C_4^p$ (i.e. \emph{PetalLength}$]4.85;+inf[$, \emph{PetalWidth}$]1.65;+inf[$, et  \emph{Class}$\{virginica\}$),
\item  un cluster ($C_3^u$) de 49 instances contenant les fleurs  moyennes versicolor caractérisées par  $C_6^p$ (i.e.  \emph{PetalLength}$]2.4;4.85]$, \emph{PetalWidth}$]0.8;1.65]$, et \emph{Class}$\{versicolor\}$).  \end{itemize}

On remarque que les variables  \emph{Class}, \emph{PetalLength}, et \emph{PetalWidth} sont fortement corrélées et les plus informatives vis-à-vis des clusters d'instances.

Pour la base CensusIncome, composée de $299.285$ instances, $11.945.874$ observations, $8$  variables numériques et $34$ variables catégorielles, les tailles de partitions de départ sont de $2$ à $128$, par puissance de $2$. Le meilleur modèle est trouvé à partir de la solution initiale correspondant à $64$ parties par variable. Le modèle post-optimisé contient  $256$ parties de variables, $607$ clusters d'instances, et $97$ clusters de parties de variables. En première analyse, notre modèle de co-clustering permet de distinguer globalement deux familles d'instances (figure~\ref{CoclustersCensus2}), les individus actifs (payeurs d'impôts, âgés de $27$ à $64$, gagnant plus que $50K$ par an, $\ldots$) et les individus inactifs (non payeurs d'impôts, âgés de moins de $15$ ans,   gagnant moins de $50K$ par an, $\ldots$).  

Globalement, le modèle obtenu permet d'obtenir un résumé de la base de données très riche en informations et exploitable à plusieurs niveaux de granularité pour piloter l'analyse exploratoire. 

\section{Conclusion}\label{Conclusion}
Dans ce papier, nous avons proposé un modèle de co-clustering des données mixtes, un critère de sélection du meilleur modèle et un algorithme d'optimisation. Nous avons montré  l'efficacité  de ce modèle pour extraire des motifs intéressants à partir des bases  petites et simples comme Iris et des bases grandes et complexes comme CensusIncome.

Toutefois, quand les données sont volumineuses et de grande complexité, notre modèle capture cette complexité et fourni un co-clustering très détaillé, au détriment de l'interprétabilité. Dans des travaux futurs, nous viserons à développer  une méthodologie permettant d'interpréter les résultats sur différents niveaux de granularité et de définir les instances et parties de variables les plus représentatives de chaque cluster pour faciliter l'interprétation du modèle. 

\bibliographystyle{rnti}
\bibliography{Biblio42}

\end{document}